\documentclass[conference]{IEEEtran}
\IEEEoverridecommandlockouts
\usepackage{cite}
\usepackage{amsmath,amssymb,amsfonts}
\usepackage{algorithmic}
\usepackage{graphicx}
\usepackage{textcomp}
\usepackage{xcolor}
\def\BibTeX{{\rm B\kern-.05em{\sc i\kern-.025em b}\kern-.08em
    T\kern-.1667em\lower.7ex\hbox{E}\kern-.125emX}}
\begin{document}

\title{Continuous Examination by Automatic Quiz Assessment Using Spiral Codes and Image Processing
\thanks{Authors acknowledge the funding of the Swedish Research Council (VR), Innovation Agency (Vinnova) and Knowledge Foundation (KKS).}
}

\author{\IEEEauthorblockN{Fernando Alonso-Fernandez, Josef Bigun}
\IEEEauthorblockA{\textit{School of Information Technology (ITE)} \\
\textit{Halmstad University}\\
Halmstad, Sweden \\
feralo@hh.se, josef.bigun@hh.se}
}

\maketitle

\begin{abstract}
We describe a technical solution implemented at Halmstad University to automatise assessment and reporting of results of paper-based quiz exams. Paper quizzes are affordable and within reach of campus education in classrooms. Offering and taking them is accepted as they cause fewer issues with reliability and democratic access, e.g. a large number of students can take them without a trusted mobile device, internet, or battery. By contrast, correction of the quiz is a considerable obstacle. We suggest mitigating the issue by a novel image processing technique using harmonic spirals that aligns answer sheets in sub-pixel accuracy to read student identity and answers and to email results within minutes, all fully automatically. Using the described method, we carry out regular weekly examinations in two master courses at the mentioned centre without a significant workload increase. The employed solution also enables us to assign a unique identifier to each quiz (e.g. week 1, week 2…) while allowing us to have an individualised quiz for each student. 
\end{abstract}


\begin{figure}[t]
\centering
\includegraphics[width=0.45\textwidth]{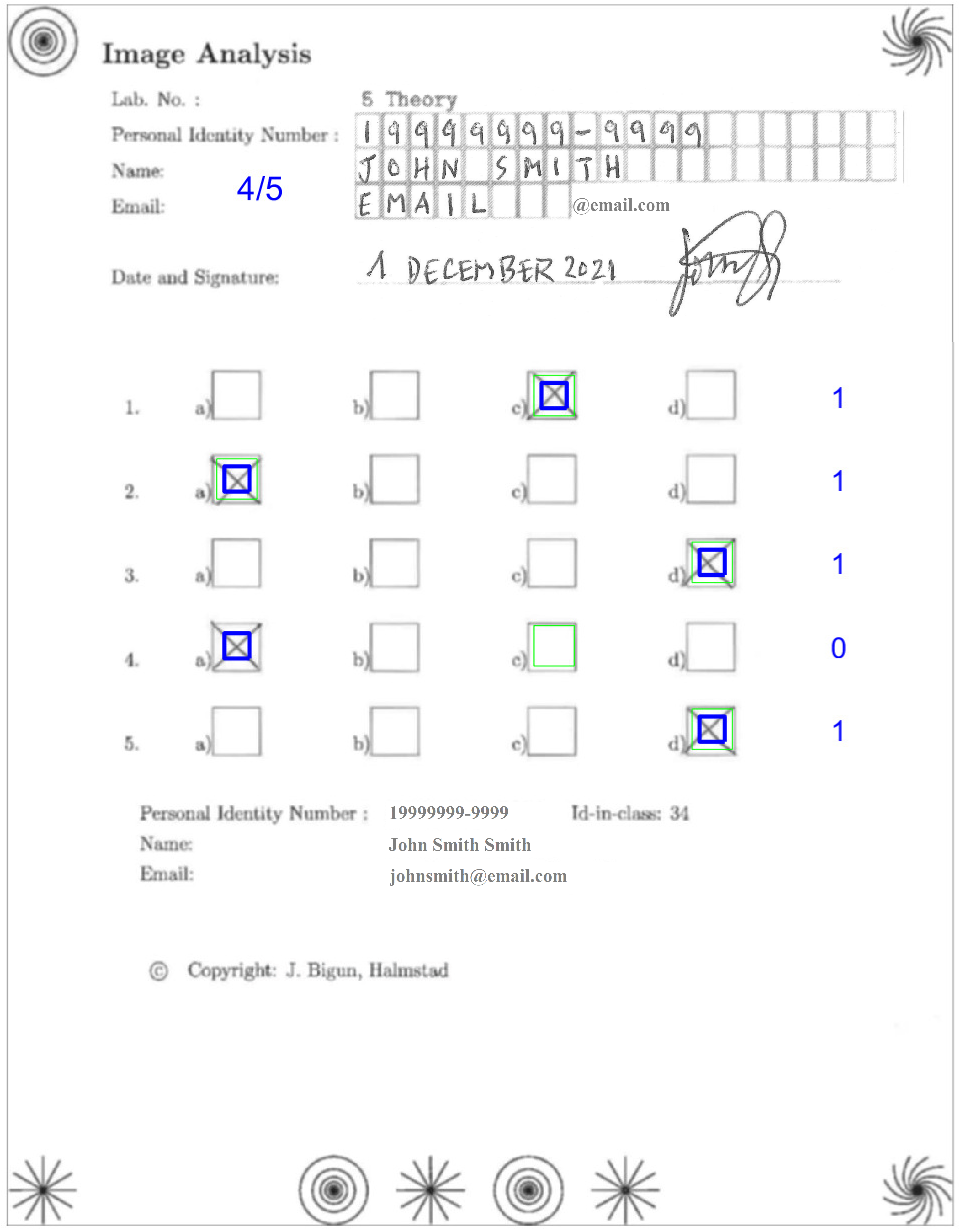}
\caption{Answer sheet with spiral codes on the corners and at the bottom.}
\label{fig:answer_sheet}
\end{figure}

\section{Introduction}

Exams or tests are a vital part of teaching, serving several goals. 
Traditionally they are seen as proof of student knowledge in a subject. However, they can also be a tool for students as well as tutors to organise studies to increase the efficiency of learning \cite{toolsforteaching}.
%
%

Among the different types of exams, multiple-choice quizzes are a popular choice to test the course understanding and breadth of students' learning~\cite{toolsforteaching}. 
Quiz questions can be difficult to construct and maintain, not because of the question and the only correct answer, but because of false alternatives.
On the one hand, false alternatives should not be too obviously false, so only students who have learned the matter sufficiently deeply are able to discern \cite{doi.org/10.3758/s13421-014-0452-8}. 
In addition, false alternatives are several times more numerous than correct alternatives, which also demands `creativity' from the teacher.
This creative effort is further magnified if, at the same time, there is a need to deliver several exam occasions per year.
Moreover, there is the need of correcting the quizzes after the examination and deliver the results to students, especially if quizzes are done often, e.g. weekly.

Therefore, efforts to reduce the burden on the preparation and correction of quizzes are desirable. 
To facilitate preparation,
quizzes can also be sequestered, i.e. the questions or the solutions are not released publicly, allowing some degree of reutilization. 
%
%
However, even sequestered, quiz questions are not too difficult to obtain
by students who have taken quizzes, for example by co-operation 
with students from previous years, or by errors made by exam
invigilators believing that quiz questions can be
made public as other (non-quiz) exams.

One goal of our solution is to automatise quiz correction, facilitating continuous examination during the course, i.e. making quizzes to be a feedback tool.
This is done by using answer sheets that contain specific spiral codes on the corners and at the bottom of the page (Figure~\ref{fig:answer_sheet}). This procedure allows to automatically: 

\begin{enumerate}
  \item align the page by detecting the spirals
  \item identify the specific quiz by using a unique combination of four spirals per quiz in the corners
  \item identify the student by using a unique combination of four spirals per student at the bottom
  \item read the student answers from the boxes since the boxes are situated at predefined positions of the page
  \item assess the correctness of questions and compute the grade
  \item deliver results to students by email
\end{enumerate}

%

The steps of the process, which will be described throughout this paper, 
are summarized in Table~\ref{tab:overview}. 
The quizzes of all students are scanned together using a multi-sheet scanner, then they are corrected and results emailed automatically to students. The entire process can take less than 30 minutes for a group of 40-50 students. 
%

%
Continuous exams increase students’ motivation for continuous study during the course since regular feedback about their learning is provided.
This is done by conducting frequent quizzes \cite{toolsforteaching} (once a week, optional), covering the content taught in the previous week. The results are saved for the final grade, so if a student passes all weekly quizzes, s/he can even skip the ordinary exam at the end of the course entirely. This allows students to reduce the stress of examinations voluntarily. In case weekly exams are not passed to a sufficient degree, timely received results become feedback and a preparation tool to improve student learning before the regular exam.
%
%
The teacher also benefits as there is more timely information on what parts have not been learned well, putting possibly one more lecture on the topic already before the exam.
The discussed technical solution about spiral generation and detection is also a topic of one of the courses where this solution is implemented. Thus, students have first-hand opportunity to observe the practical application of a concept taught in the course as well.

The rest of the paper is organised as follows.
In Section~\ref{sect:spiralcodes},
we describe the algorithm used to generate and detect the spiral codes.
Its specific implementation to automatise quiz correction and enable quiz personalization is described in Section~\ref{sect:correction-personalization}.
In Section~\ref{sect:continuous-examination},
we discuss the employed solution from the point of view of continuous examination and its impact on the students' learning.
Finally, conclusions are given in Section~\ref{sect:conclusions}.


\begin{table}[t]
\caption{Automatic quiz correction.}
\begin{center}
\begin{tabular}{c}

\includegraphics[width=0.44\textwidth]{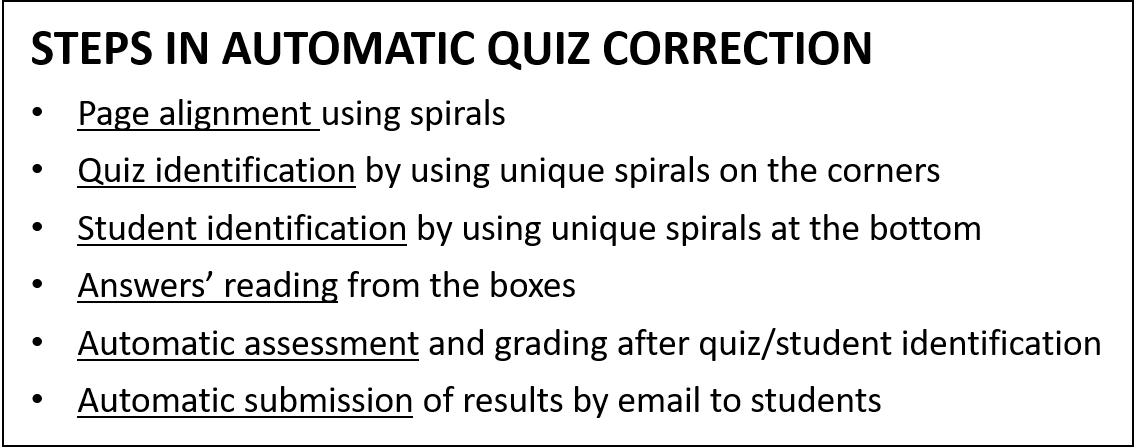}

\end{tabular}
\label{tab:overview}
\end{center}
\end{table}

\section{Spiral Codes}
\label{sect:spiralcodes}

\begin{figure}[htb]
\centering
\includegraphics[width=0.44\textwidth]{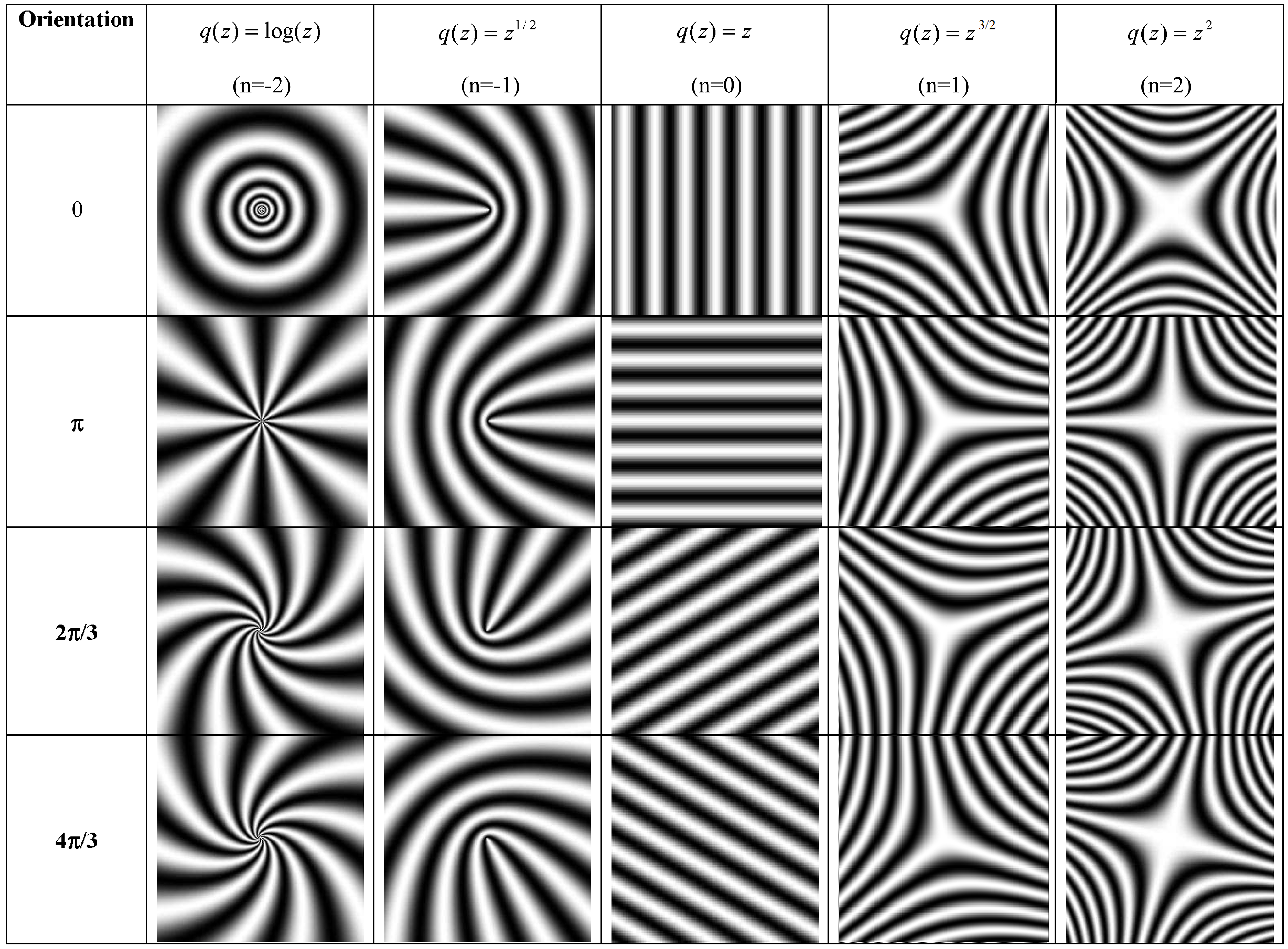}
\caption{Example of symmetric patterns. Each column (from 2 to 6)
represents one family of patterns (defined by the value of $n$ in Equation~\ref{eq:qfunction}) differing only by their group orientation 2$\varphi$ given in column 1.}
\label{fig:symmetric_patterns}
\end{figure}

Symmetry features are complex-valued and enable the detection and description
of symmetric patterns of spirals such as lines, circles, parabolas, stars, etc. in
a local neighbourhood  (Figure~\ref{fig:symmetric_patterns}). 
The description of the pattern family is done by estimation
of the orientation simultaneously with the detection of the pattern
itself. 
These features are extracted via symmetry filters, which output how much of a certain type exists in a local image neighbourhood.
This section describes the process of generating and detecting the mentioned patterns. 
We provide further details of this method at \cite{[Bigun04],[Bigun06]}.

\subsection{Synthesis of spirals}

Symmetric patterns can be generated by harmonic function pairs \cite{BIGUN1997290}. 
Let
$\xi \left( {x,y} \right)$ and $\eta \left( {x,y} \right)$ be harmonic functions satisfying the Cauchy-Reimann equations:

\begin{equation}
\begin{array}{*{20}{c}}
{\frac{{\partial \xi }}{{\partial x}} = \frac{{\partial \eta }}{{\partial y}}}&{{\rm{and}}}&{\frac{{\partial \xi }}{{\partial y}} = \frac{{\partial \eta }}{{\partial x}}}
\end{array}
\end{equation}

In other words, the iso-curves of $\xi$ and $\eta$ are orthogonal to each other. Then $\xi$ and $\eta$ are said to be harmonic function pairs (HFPs). These harmonic function pairs can be easily obtained as the real and imaginary parts of any analytic
function $q(z)$. However, if $q$ in particular fulfills $\frac{{dq}}{{dz}} = {z^{\frac{n}{2}}}$ with $z = x+iy$ and $n$ being integer, then polynomial expansion of the local orientation of any pattern can be achieved. The analytic function $q(z)$ will then be defined as

\begin{equation}
q\left( z \right) = \left\{ {\begin{array}{*{20}{c}}
{\frac{1}{{\frac{n}{2} + 1}}{z^{\frac{n}{2} + 1}}}&{{\rm{if}}}&{n \ne  - 2}\\
{\log \left( z \right)}&{{\rm{if}}}&{n =  - 2}
\end{array}} \right.
\label{eq:qfunction}
\end{equation}

For example, if $q(z)=\log(z) $  then:

\begin{equation}
q(z)=\log(z) = \log(r) +i\theta
\end{equation}

%
%
%

\begin{equation}
\xi \left( {x,y} \right) = \Re \left( {q\left( z \right)} \right) = \log(r)
\end{equation}

\begin{equation}
\eta \left( {x,y} \right) = \Im \left( {q\left( z \right)} \right) = \theta
\end{equation}

\noindent with $\Re$ and $\Im$ representing the real and imaginary parts, whereas $r$ and $\theta$ are the magnitude and arguments of $z$, respectively. 
The linear combination of $\xi$ and $\eta$ generates a set of patterns belonging to the same family (generated by $\xi$ and $\eta$). For illustration, the symmetric patterns generated by using the 1D function $f$ of Equation~\ref{eq:functionf} is shown in Figure~\ref{fig:symmetric_patterns}, with parameters $\left( {a,b} \right) = \left( {\cos \varphi ,\sin \varphi } \right)$.

\begin{equation}
\label{eq:functionf}
f\left( {\xi ,\eta } \right) = \cos \left( {a\xi  + b\eta } \right) + 1
\end{equation}

\begin{equation}
f\left( {\xi ,\eta } \right) = \cos \left( {a\Re \left( {q\left( z \right)} \right) + b\Im \left( {q\left( z \right)} \right)} \right) + 1
\end{equation}

The pattern family generated by $q(z)=\log(z)$ 
%
%
in Figure~\ref{fig:symmetric_patterns} (second column)
represent the pattern of `bended spokes' (except for $\varphi=0$, which corresponds to concentric circles),
%
%
while other families represent parallel lines, parabolas, stars, etc.
%
%
%
%
%
In every pattern family, members are distinguished from each other by the angle parameter $\varphi$. The parameter has a precise meaning deducible from the pattern itself and which an image processing algorithm can extract. In the family of $q\left( z \right) = \log \left( z \right)$,
%
%
the angle $\varphi$ (except for $\varphi=0$) represents the `spirality' of the iso-curves (i.e. bending of the spokes), also called chirality. In all other families, the angle $\varphi$ identifies the amount of rotation of the pattern itself.

\begin{figure}[t]
\centering
\includegraphics[width=0.44\textwidth]{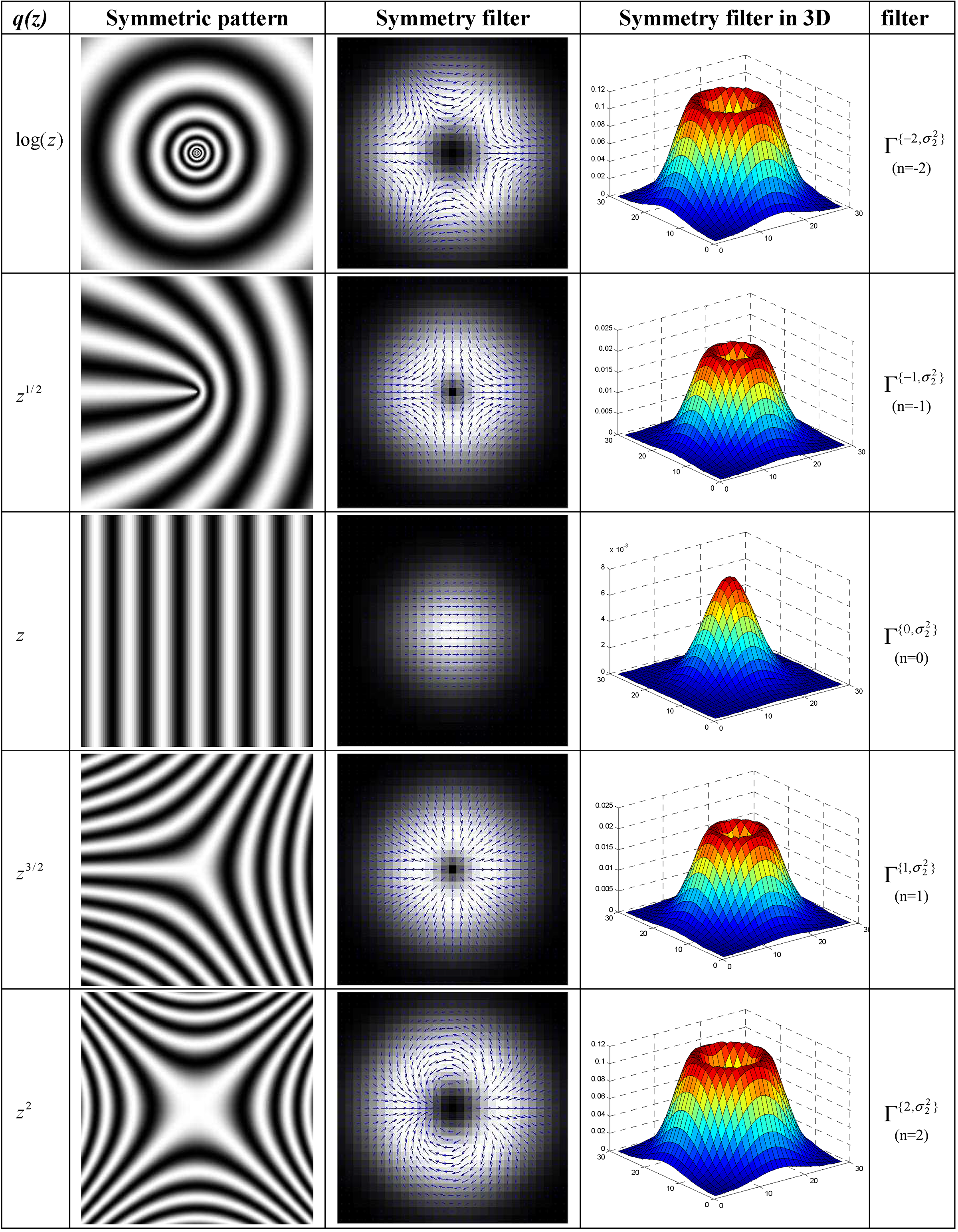}
\caption{Symmetric patterns and their associated detection filters.}
\label{fig:symmetry_filters}
\end{figure}

\begin{figure*}[htb]
\centering
\includegraphics[width=0.9\textwidth]{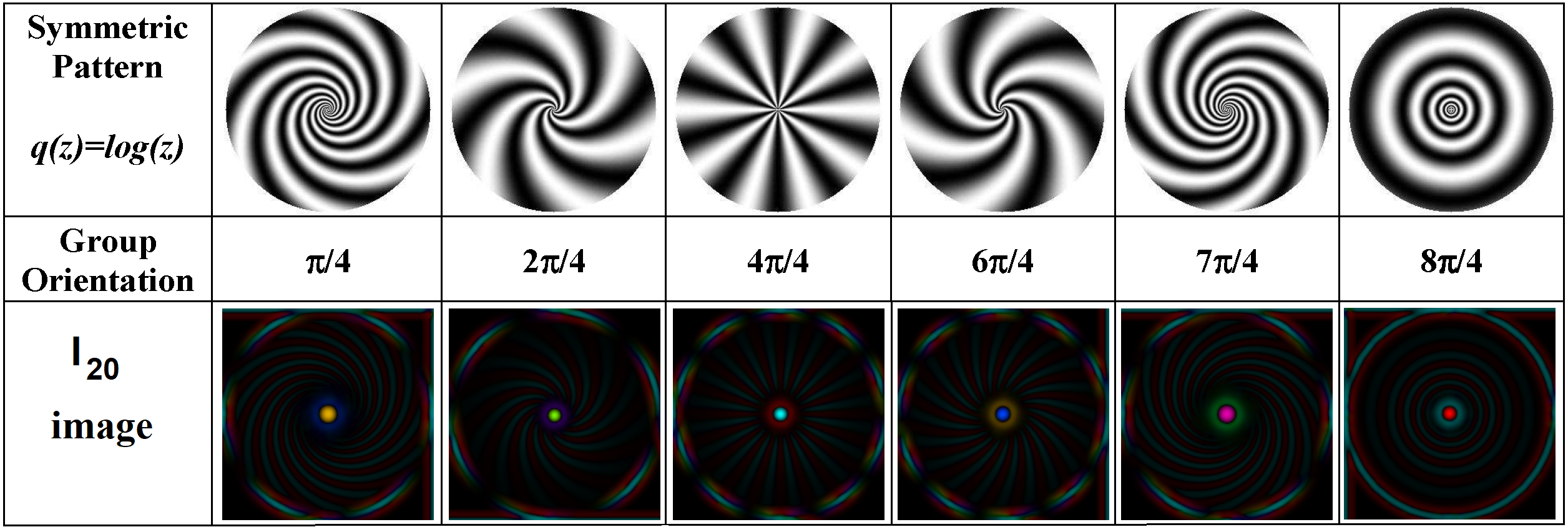}
\caption{Encoding quiz identity. The image of the third row shows the $I_{20}$ image of the patterns of the first row. Note that $I_{20}$ is a complex-valued matrix whose angle is used as the hue for the image shown in the third row, and the saturation represents the complex magnitude. To depict the magnitude, the values are re-scaled, so the maximum intensity represents the maximum magnitude, while black represents zero magnitude. Zero angle is given by red colour (which corresponds to the pattern of concentric circles, last column). Images of the last row are not thresholded, magnitude is at least three times larger in the centre than elsewhere. Group orientation equals to $2\varphi$.}
\label{fig:spirals_identity1}
\end{figure*}

\subsection{Symmetry filters and detection of symmetric patterns}

Symmetry filters are a family of filters computed from symmetry
derivatives of Gaussians. They are used by the image processing software to detect and to identify the member of a symmetric family pattern. The $n^{th}$ symmetry derivative of a
a 2D Gaussian function, $\Gamma ^{\left\{ {n,\sigma ^2 } \right\}}$, is obtained
by applying $n$ times the partial derivative operator $D_x  + iD_y  = \left(
{\partial /\partial x} \right) + i\left( {\partial /\partial y}
\right)$, called 1$^{st}$ symmetry derivative, to a 2D Gaussian $g(x,y)$:

\begin{equation}
\Gamma ^{\left\{ {n,\sigma ^2 } \right\}}  = \left\{
{\begin{array}{*{20}c}
   {\left( {D_x  + iD_y } \right)^n g(x,y)} & {\left( {n \ge 0} \right)}  \\
   {\left( {D_x  - iD_y } \right)^{\left| n \right|} g(x,y)} & {\left( {n < 0} \right)}  \\
\end{array}} \right.
\label{eq:nsymmder}
\end{equation}

Since $D_x  + iD_y$ and $\left( { - \frac{1}{{\sigma ^2 }}}
\right)\left( {x + iy} \right)$ behave identically when acting on a
Gaussian \cite{[Bigun04],[Bigun06]}, 
Equation~\ref{eq:nsymmder} can
be rewritten as:

\begin{equation}
\Gamma ^{\left\{ {n,\sigma ^2 } \right\}}  = \left\{
{\begin{array}{*{20}c}
   {\left( { - \frac{1}{{\sigma ^2 }}} \right)^n \left( {x + iy} \right)^n g(x,y)} & {\left( {n \ge 0} \right)}  \\
   {\left( { - \frac{1}{{\sigma ^2 }}} \right)^{\left| n \right|} \left( {x - iy} \right)^{\left| n \right|} g(x,y)} & {\left( {n < 0} \right)}  \\
\end{array}} \right.
\label{eq:nsymmder_alt}
\end{equation}

%
It has been shown that these symmetry derivatives of Gaussians
are able to detect patterns as those of Figure~\ref{fig:symmetric_patterns} efficiently through complex convolutions interleaved with pixel-wise applied complex squaring:

\begin{equation}
I_{20} = \left\langle {\Gamma ^{\left\{ {n,\sigma _2^2 } \right\}}
,h} \right\rangle \label{eq:I20}
\end{equation}

\noindent where $h$ is the complex-valued orientation tensor field
given by:

\begin{equation}
h = \left\langle {\Gamma ^{\left\{ {1,\sigma _1^2 } \right\}} ,f}
\right\rangle ^2 \label{eq:I20linear}
\end{equation}

\noindent and $f$ is the image under analysis \cite{[Bigun06]}.
Parameter $\sigma_1$ defines the size of the derivation filters used
in the computation of image $h$, whereas $\sigma_2$, used in the
computation of $I_{20}$, defines the size extension of the sought
pattern.


For each family of symmetric patterns, there is a symmetry filter
$\Gamma ^{\left\{ {n,\sigma ^2 } \right\}}$ (indexed by $n$)
suitable to detect the whole family \cite{[Bigun97a]}, as well as to identify family members (by estimating $\varphi$). 
Figure~\ref{fig:symmetry_filters} indicates the filters that are used to
detect each family. The local maxima in $\left| {I_{20} } \right|$
give the location, whereas the argument of $I_{20}$ at maxima
locations 
identifies the member of the family (the angle $2\varphi$, due to squaring in Equation~\ref{eq:I20linear}) associated with the filter (index $n$). 
%
%
Therefore, $I_{20}$ jointly contains the position and orientation of the pattern,
encoding how much of a certain type of symmetry 
exists in a local neighbourhood  of the image $f$. Thus, a
single symmetry filter $\Gamma ^{\left\{ {n,\sigma ^2 } \right\}}$
is used for the recognition of the entire family of patterns,
regardless of their orientation (or chirality). 
We have applied symmetry filters successfully to a wide range of detection tasks
such as cross-markers in vehicle crash tests \cite{[Bigun91]},
core-points and minutiae in fingerprints \cite{nilsson03prl,[Fronthaler08]}, 
or eye region on face images \cite{[Alonso15]}. 
Inherent to Equations~\ref{eq:I20}-\ref{eq:I20linear} is the fact that
%
%
$I_{20}$ is computed by filtering in Cartesian
coordinates without the need for transformation to curvilinear
coordinates (which is implicitly encoded in the filter).

The choice of the spiral family $q\left( z \right) = \log \left( z \right)$ to encode information via its angle parameter $\varphi$ is motivated by the fact that this angle does not correspond to a geometric rotation of the pattern itself. Thereby, the parameter of a member is independent of rotation as well as the scale of the (spiral) pattern. This affords image processing to identify and locate family members accurately independent of rotation and despite large scale variations of the pattern.

\section{Automatic Correction and Quiz Personalization}
\label{sect:correction-personalization}

The quiz is made on paper using the template shown in Figure~\ref{fig:answer_sheet} as answer sheet. Four spiral codes are placed in the corners and four on the bottom, which serve to automatise the correction process once the sheets are scanned. 
The spirals can be produced with mathematical precision and detected with sub-pixel accuracy by searching for the local maxima of $|I_{20}|$, as described in the previous section. At the same time, the argument of $I_{20}$ at the local maxima will decode the exact member of the family.
Figure~\ref{fig:spirals_identity1} shows different members of the family $q\left( z \right) = \log \left( z \right)$ in the first row ($n$=-2), and the resulting $I_{20}$ image after applying the corresponding filter $\Gamma ^{\left\{ {-2,\sigma ^2 } \right\}}$.
A maximum is clearly appreciable at the centre of the patterns, which allows their localisation.
The hue in the $I_{20}$ image is used to represent the argument, which, as it can be seen, it is different for each member of the (decoding) family. This allows its simultaneous identification as well.
Reliable localisation and identification at the same time is unique to spiral codes, as nearly 50\% of every spiral can be allowed to be affected by noise or even occluded without affecting the result. 
This is because the redundancy of the isocurves in the patterns optimizes the ability of symmetry filters to detect and identify them even when part of the image is occluded or degraded.

By detecting the spirals, we can precisely align the page thanks to rotation and scale invariance of $\log \left( z \right)$ spirals. This allows identifying the various fields filled by the student, including name, ID number, and answers since the position of the corresponding boxes is predefined. 
%
The answer to each question can be easily read since unfilled boxes will have their pixels close to white, while filled boxes will have a certain percentage of pixels with a darker value.
%
%
On the top left corner, the spiral of concentric filters is always used as a way to identify the corner if the paper happens to be rotated. 
This allows to 
correct the alignment if, for example, the image is scanned upside down.
The other three corners carry spirals with varying chirality to identify the course, the specific subject inside the course, and the question set in the subject area, which offers hundreds of different possibilities currently.  
%
%
Using spirals from the same family also has the advantage that they can be extracted using a single symmetry filter, via Equation~\ref{eq:nsymmder_alt}. 
Since the specific spirals used in the corners allows to identify the quiz question set, this, in turn, defines which is the correct answer for each question.

The four spirals at the bottom in Figure~\ref{fig:answer_sheet} identify the student, i.e. the name, email, and personal identity number, which are also printed on the answer sheet (John Smith, etc.). This is because, for each student, we produce a different quiz set drawn from a pool of questions. The signature, name, email, and dates written by hand at the top generally do not influence automatic processing. They serve only for non-repudiation, confirmation by the student that the printed information is indeed correct, or whether this information needs correction. The four spiral codes at the bottom, when decoded, corresponds to the printed student information. 

An important aspect is that we can choose different combinations of spirals 
to create unique exams or to number different pages of answer sheets. 
By using different combinations on the corners, we identify 
%
the quiz itself (e.g. course `A', week `X', laboratory part, page `B'...) at no additional computational cost, 
enabling the automatic correction. 
For example, if 
we use the six members shown in Figure~\ref{fig:spirals_identity1} and reserve the pattern of concentric circles for the top left corner only (to facilitate unique identification of this corner),
in the other three corners, we can have up to 
%
$5^3=125$ variations of spirals (ordered and with repetition).
This number could be made higher easily by using more than four spirals. 
For example, an extra spiral chosen among the five available in Figure~\ref{fig:spirals_identity1} would give $5^4=625$ variations, and using all would give $5^5=3125$ variations.
Spirals from other families could be used as well for even more possibilities.
This would, however, demand the use of more than one symmetry filter for detection, but with today's computing capabilities, it would not add too much execution overhead.
%
%
%
%
A few student identity spirals (shown at the bottom) also allow producing an individual quiz for each student, so each student can get his/her own set of questions.
Here, the same considerations regarding the different variations also apply. If we use the six spirals of Figure~\ref{fig:spirals_identity1} in groups of four at the bottom of the page, this would allow up to $6^4=1296$ individual quizzes.

\begin{figure*}[t]
\centering
\includegraphics[width=0.9\textwidth]{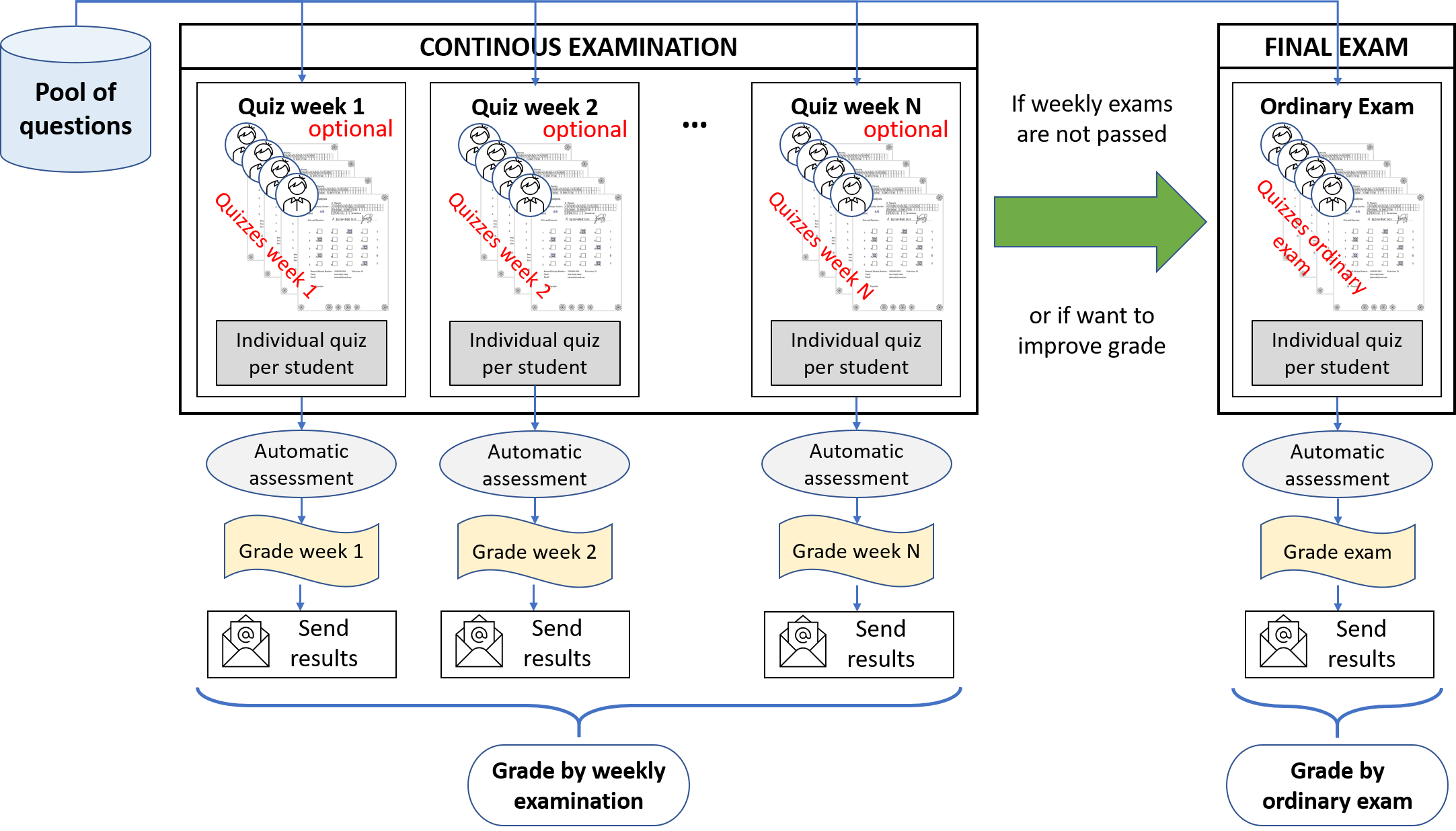}
\caption{Pipeline of the employed continuous examination solution with automatic correction and quiz personalization.}
\label{fig:solution_pipeline}
\end{figure*}

Therefore, after the mentioned process, we obtain from each answer sheet automatically:

\begin{itemize}
  \item The ID of the student 
  \item The ID of the quiz itself 
  \item The answers to the quiz 
\end{itemize}

As shown in Figure~\ref{fig:answer_sheet}, assessment of student answers is done 
thanks to accurate alignment, quiz identification, and student identification procedures afforded by the spiral codes. The alignment and quiz identification are made together via four spirals at the corners, whereas student identity recognition is achieved via four spirals at the bottom. 
The coloured data represents the assessment. Green and blue in boxes show where correct answers and student answers are, respectively, whereas numeric data in blue are assessments. The system then delivers the assessment result by email at the detail level decided by the teacher policy to students (e.g. overall grade only, or question by question). Students are, however, entitled to see the assessment on a computer screen in full detail upon appointment.

Alignment is a critical step for automatic assessment. Such step corrects the image precisely and robustly via automatic geometric transformation (rotation, translation, and scaling). However, because spirals contain redundant information while being rotation and scale-invariant, robust localisation of spiral centres at sub-pixel precision is possible reliably by image processing. If one part of a spiral is noisy or even occluded completely, the remainder of it has still powerful information to locate the centre precisely. Figure~\ref{fig:answer_sheet} shows the image after automatic alignment where all corner spirals touch sheet borders precisely, as they should. The complex convolution procedure we suggest not only localises spirals precisely but also identifies spirals accurately at negligible implementation and computation cost. 

The student identity decoding is actually not a new system, conceptually. It uses the same spiral detection and identification procedure used for alignment and quiz identity recognition just described. Consequently, 100\% correct spiral identification and precise localisation allowing automatic assessment have been observed on tens of thousands of spirals in images similar to Figure~\ref{fig:answer_sheet}, as the automatic assessment is also shown to teacher while it progresses.

While integrity/cheating is, in general, a challenging problem for any exam, in groups of 50-60 students participating in our quizzes, it is even more manageable in some respects. Each student gets her/his own private quiz so that it does not help to look at the results of the fellow student sitting next. The fact that weekly quizzes are voluntary and the amount of time is just enough ($\approx$1 minute per question) to identify the right answer also helps students to focus on their quizzes and learning. One drawback of the student-specific quizzes on paper is that they add a slight overhead to distribute them to their owners, which is a logistics problem. We have addressed this by putting quiz papers at 4-5 different piles and letting fellow students distribute quizzes to their owner students lining up, as each quiz has already student name details printed on it. 

Evidently, the latter can be abused, just as in any exam, if it is not accompanied by invigilation. Currently, we do this by asking students to put their photo id-cards on the table while seated to answer the quiz. The teacher verifies identities while students take the quiz, with minimal disturbance to students. However, the present system offers potential in that regard. Invigilation can be reinforced with biometric identification. Students sign the quiz sheet for legal reasons as illustrated in Figure~\ref{fig:answer_sheet}, and we are able to find the location where the signature is expected thanks to precise alignment. Accordingly, biometric signature verification can be easily integrated \cite{[Gilperez08]}. However, because social and privacy aspects of this must also be studied in the integration, we have not yet implemented this potential reinforcement.

\begin{figure*}[t]
\centering
\includegraphics[width=0.95\textwidth]{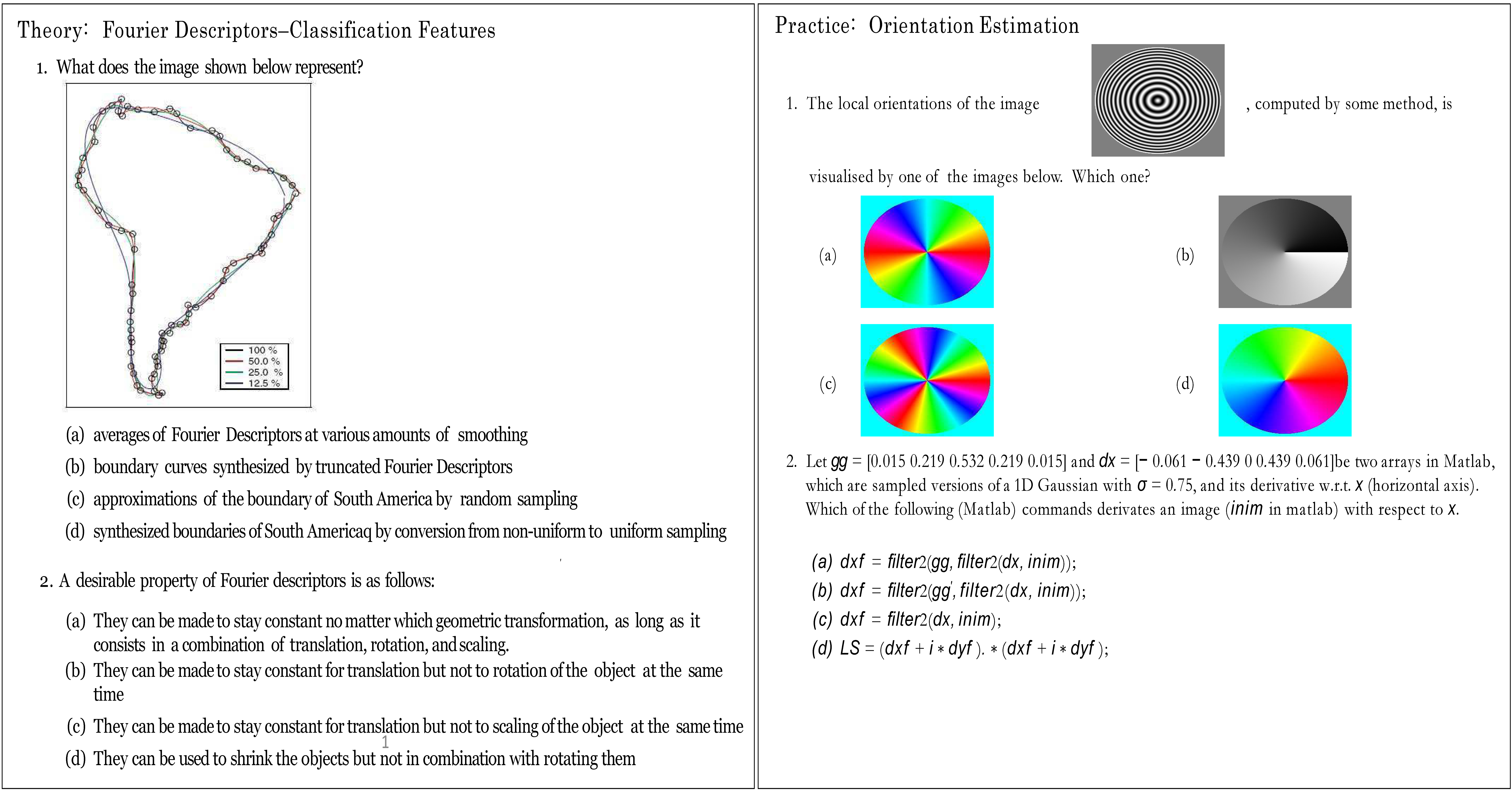}
\caption{Quiz question examples (left: theory, right: practice).}
\label{fig:example_question}
\end{figure*}

\section{Continuous Examination and Impact on Students' Learning}
\label{sect:continuous-examination}

The employed solution allows frequent and regular quiz examinations. 
The pipeline of this solution is shown in Figure~\ref{fig:solution_pipeline}.
Quiz results are calculated and emailed automatically to the students as soon as they are scanned, as correction is within minutes. This is currently practised in two master courses at Halmstad University: Image Analysis (6 weeks, $\sim$45 students) and Computer Vision in 3D (5 weeks, $\sim$20 students).
One multiple-choice quiz per week with 20 questions is carried out concerning only subjects treated during the previous week, which corresponds to 4 hours of lectures and 4 hours of laboratory. The quiz is held for 15 minutes at the first lecture of each week, with half of the questions related to theory and the other half to practice.
For each course, we keep one pool of potential quiz questions with four answer alternatives built across the years. 
The pool currently has ca 300 questions per course, corresponding to the different parts of the course, with approximately 60 questions per teaching of 1 week. 
For weekly quizzes, questions are selected randomly among one half of the pool for each topic, while the other half is reserved for the ordinary exam at the end of the course. This is to ensure that the chosen questions are different in each exam. 
This also allows certain flexibility for the purpose of not having exactly the same quiz for a student on the same topic. 
To reduce the leaking of questions from the pool as much as possible, we also apply a policy of not making the exam questions or solutions public and not allowing that the exam questionnaire is taken home by the student after the exam. This is practicable as the entire quiz is recollected. However, spiral codes afford to place alternative boxes right next to the question itself, a potential yet to be exploited. In Figure~\ref{fig:example_question}, some examples of quiz questions are shown.

Weekly quizzes are offered as optional. The ordinary exam at the end of the course is also a quiz, but a larger one, covering all weeks of the course. If the results of weekly quizzes are not good for Pass degree, the student can use the received feedback to prepare for the ordinary exam more efficiently. Currently, the grade of each weekly quiz is saved, so if a student is satisfied with the result of a week, s/he may not have to do the ordinary exam to pass the course. Students are also allowed to try to improve the results of weekly quizzes, if already sufficient for Pass, in the ordinary exam, without risk of grade degradation (i.e., the highest grade is used). 

Thereby, the ordinary exam itself becomes optional if the students pass each weekly quiz. In practice, ca 25\% of the students do the ordinary exam, with the rest having received their final grade via weekly quizzes conducted throughout the teaching term. Continuous examination becomes a learning tool rather than a control tool, which motivates the students to study regularly during the course while receiving continuous feedback.

Outcomes of the described solution have been so far satisfactory after several years of practice. Students enhance their academic education by being motivated to continuous study, rather than preparing for the ordinary exam `in the last minute'. They also receive regular and continuous feedback about their learning throughout the entire course. 
Results of anonymous opinion polls carried out at the University among course students also show an excellent degree of satisfaction by the students with the two courses. Although there is no specific question related to continuous exams, most students go an extra length in free-text comments to explicitly praise it as a significant contributor to their learning.
An additional benefit is that it illustrates the practice of image processing, the very knowledge that is learned.

\section{Discussion and Conclusions}
\label{sect:conclusions}

Quiz exams demand a significant time investment, both in producing questions on every new exam occasion and in assessing them afterwards.
Here, we present a technical solution to automatise assessment and reporting of results of paper-based quizzes. 
The described method is able to obtain student identity and answers from the answer sheets and email the results afterwards, all in an automatic fashion. 
%
%
The only manual intervention happens during quiz scanning, which can be done in minutes thanks to the use of a multi-sheet scanner. The employed scanner produces a single pdf document containing all the answer sheets. With the described solution, we are able to discern the student to which each answer sheet corresponds, read the answers to the quiz, and communicate the results.
%
%
%

Thanks to the described solution, we carry out regular weekly examinations of the concepts treated the previous week in two engineering master courses at Halmstad University: Image Analysis and Computer Vision in 3D. 
The quizzes are optional, but if they are passed, the corresponding part in the ordinary exam at the end of the teaching term can be skipped. This enables continuous examination during the courses, contributing to keeping students motivated by studying regularly and receiving regular feedback. 
Students can also drop their lowest score by repeating that part in the ordinary exam and retaining the highest score of the two. This option is available only for those who want to pass the course at minimum Pass grade. 
If the weekly quizzes are not passed, students can still use the feedback received to improve their results in the ordinary exam. 
All in all, this solution gives students a greater opportunity to do well in the course \cite{toolsforteaching}.
A proof of the success of this solution is that only 25\% of the students do the ordinary exam at the end of the course.
Results of the opinion polls carried out at the University also shows an excellent degree of satisfaction by the students.
Automatic assessment solutions also allow bias mitigation since there is very little human intervention during the correction \cite{doi:10.1177/0004944116664618}.

It is worth mentioning a limitation that became all too obvious during covid-19 restrictions. All campus education, including exams, were cancelled. Evidently, paper-based continuous exams have a `drawback' as classroom education is a prerequisite for its implementation.

%

\bibliographystyle{IEEEtran}




\end{document}